\documentclass[oneside, 11pt]{article}
\usepackage{times}

\usepackage{soul}
\usepackage{url}
\usepackage[hidelinks]{hyperref}
\usepackage[utf8]{inputenc}
\usepackage[small]{caption}
\usepackage{graphicx}
\usepackage{amsmath}
\usepackage{booktabs}
\usepackage{multirow}
\usepackage{multicol}
\usepackage{diagbox}
\usepackage{wrapfig}
\usepackage{xcolor}
\newcommand{\nosection}[1]{\vspace{2pt}\noindent\textbf{#1.}}
\begin{document}
\newcommand\todo[1]{\textcolor{red}{#1}}
\newcommand{\hengtong}[1]{\textcolor{blue}{#1}}

\title{A Survey of Trustworthy Graph Learning: Reliability, Explainability, and Privacy Protection}
%
%
\author{
\normalfont{Bingzhe Wu$^{1}$\thanks{Correspondence to: Bingzhe Wu, email: \texttt{wubingzhe94@gmail.com}}} 
\and
\normalfont{Jintang Li$^{2}$}
\and
\normalfont{Junchi Yu$^{3}$} 
\and
\normalfont{Yatao Bian$^{1}$} 
\and
\normalfont{Hengtong Zhang$^{1}$} 
\and
\normalfont{Chaochao Chen$^{4}$}
\and
\normalfont{Chengbin Hou$^{1}$}
\and
\normalfont{Guoji Fu$^{1}$}
\and
\normalfont{Liang Chen$^{2}$}
\and
\normalfont{Tingyang Xu$^{1}$} 
\and
\normalfont{Yu Rong$^{1}$} 
\and
\normalfont{Xiaolin Zheng$^{4}$}
\and
\normalfont{Junzhou Huang$^{5}$} 
\and
\normalfont{Ran He$^{3}$} 
\and
\normalfont{Baoyuan Wu$^{6}$}
\and
\normalfont{Guangyu Sun$^{7}$}\and
\normalfont{Peng Cui$^{8}$}\and
\normalfont{Zibing Zheng$^{2}$}\and
\normalfont{Zhe Liu$^{10}$} \and
\normalfont{Peilin Zhao$^{1}$} 
\and
\\
{\normalfont $^1$Tencent AI Lab}\\
$^2$Sun Yat-sen University\\
$^3$Institute of Automation, Chinese Academy of Sciences\\
$^4$Zhejiang University \\
$^5$University of Texas at Arlington\\
$^6$School of Data Science, Shenzhen Research Institute of Big Data,
\\The Chinese University of Hong Kong, Shenzhen\\
$^7$Peking University \\
$^8$Tsinghua University \\
$^{10}$Zhejiang Lab\\
[.5em]
}

%



\maketitle            
\begin{abstract}
Deep graph learning has achieved remarkable progresses in both business and scientific areas ranging from finance and e-commerce, to drug and advanced material discovery. Despite these progresses, how to ensure various deep graph learning algorithms behave in a socially responsible manner and meet regulatory compliance requirements becomes an emerging problem, especially in risk-sensitive domains. Trustworthy graph learning (TwGL) aims to solve the above problems from a technical viewpoint. In contrast to conventional graph learning research which mainly cares about model performance, TwGL considers various reliability and safety aspects of the graph learning framework including but not limited to robustness, explainability, and privacy. In this survey, we provide a comprehensive review of recent leading approaches in the TwGL field from three dimensions, namely, reliability, explainability, and privacy protection. We give a general categorization for existing work and review typical work for each category. To give further insights for TwGL research, we provide a unified view to inspect previous works and build the connection between them. We also point out some important open problems remaining to be solved in the future developments of TwGL.
\end{abstract}

\section{Introduction}
Recently, Deep Graph Learning (DGL) based on Graph Neural Networks (GNNs) have emerged as a powerful learning paradigm for graph representation learning. In the past few years, DGL is becoming an active frontier of deep learning with an exponential growth of research. With advantages in modeling graph-structured data, DGL has achieved remarkable progress in many important areas, ranging from  finance (e.g., fraud detection and credit modeling) \cite{fraud_gnn,credit_gnn, bian2020rumor,li2019semi},  e-commerce (e.g., recommendation system) \cite{fedgnn_rec,min2022masked}, drug discovery  and advanced material discovery \cite{guo2021dockstream,molecularnet,rong2020self,ma2022cross,mao2021molecular}.


Although considerable progress has been made in the field of deep graph learning, there are many challenges about the level of \emph{trustworthiness} associated with GNNs in many real-world applications, especially in risk-sensitive and and security-critical scenarios such as finance and bioinformatics \cite{rong2020self}. Such challenges are real and call for sophisticated solutions, as many untrustworthy sides of DGL models have been exposed through adversarial attacks \cite{DBLP:journals/corr/abs-2003-05730,chang2020restricted}, inherent bias or unfairness \cite{zhang2022fairness}, and lack of explainability \cite{yuan2020explainability} and privacy protection \cite{chen2020survey} in the current rapidly evolving works. In light of these challenges, various emerging laws and regulations in artificial intelligence (AI) ethic, e.g., EU General Data Protection Regulation (GDPR) \cite{parliament2016regulation}, promote different companies to develop trustworthy algorithms for meeting regulatory compliance requirements of privacy, explainability and fairness, etc. 



\begin{wrapfigure}{r}{0.5\linewidth}
  \centering
  \includegraphics[width=\linewidth]{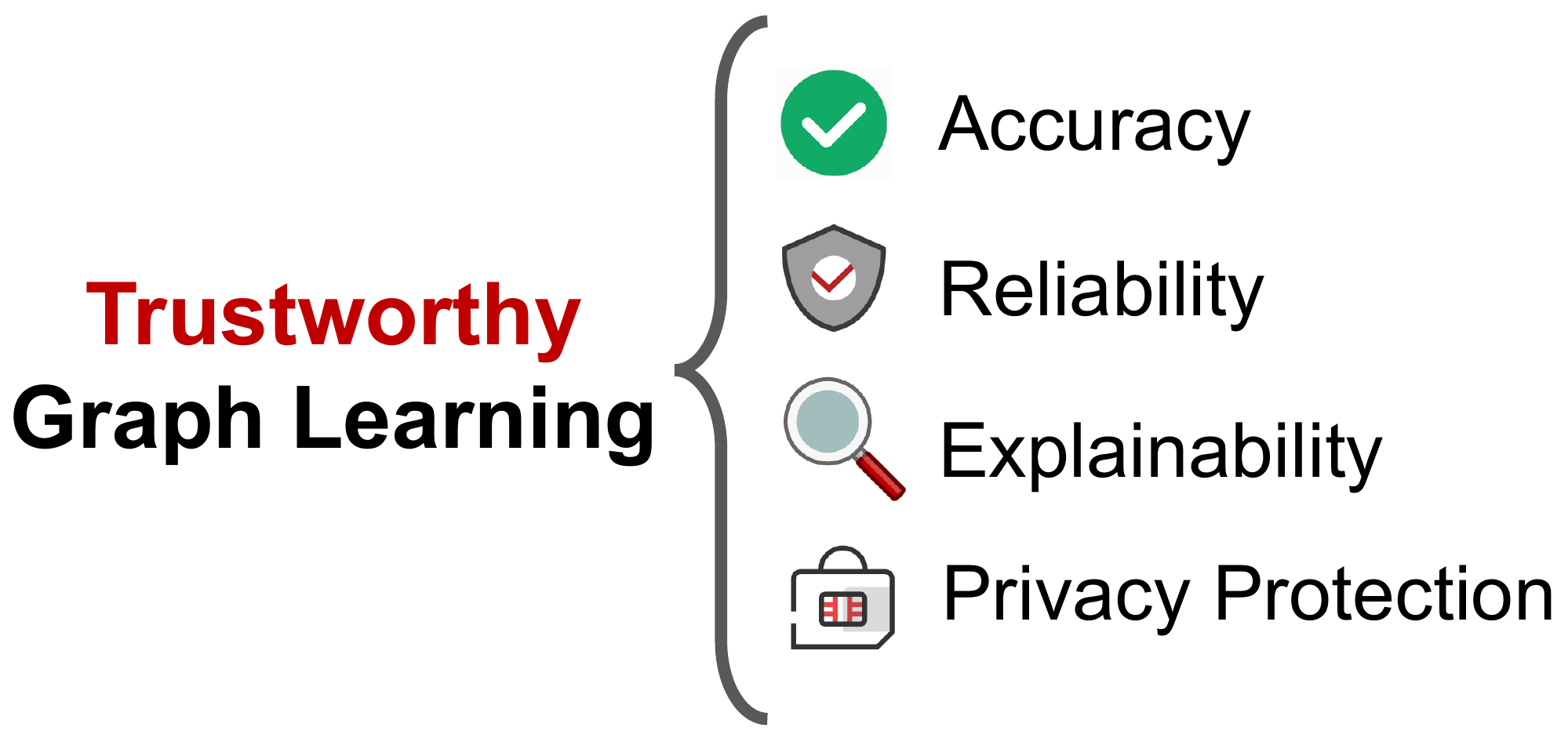}
  \caption{Four key dimensions of trustworthy graph learning.}
  \label{fig:trustworthy}
\end{wrapfigure}

A new research trend in recent years has been to investigate ways to to build trustworthy algorithms and ensure they behave in a socially responsible manner. Trustworthy Graph Learning (TwGL), an emerging field with the aim to solve the trustworthiness issues in graph data from a technical viewpoint, has become a focus of attention in the areas of deep graph learning. In contrast to conventional graph learning research which mainly focus on how to develop accurate models, TwGL considers various reliability and safety aspects of the graph learning problem beyond model performance. 
In Figure~\ref{fig:trustworthy}, we summarise the four key dimensions towards reliable TwGL, that is, \emph{accuracy}, \emph{reliability}, \emph{explainability}, and \emph{privacy protection}.

We give detailed descriptions as follows: 1) \textbf{Accuracy or utility}: The prediction accuracy is the basic ability of a trustworthy model. Trustworthy GNNs are expected to generate accurate output, consistent with the ground truth, as much as possible; 2) \textbf{Reliability}: Trustworthy GNNs  should be resilient and secure. In other words, they must be robust against different potential threats, such as inherent noise, distribution shift, and adversarial attacks; 3) \textbf{Explainability}. Understanding the graph learning model is another important aspect in developing trustworthy GNNs. The model itself must allow explainable for the prediction, which can help humans to enhance understanding, make decisions and take further actions; 4) \textbf{Privacy protection}: Trustworthy GNNs are required to ensure full privacy of the models as well as data privacy.

\nosection{Motivation and Contributions} 
The trustworthiness of DGL models is an inescapable problem when they step into real-world applications, posing various open, unexplored, and unidentified challenges in both academic and industrial communities. 
Therefore, various aspects of DGL models are required to be carefully considered to achieve trustworthiness. To facilitate the research in the field of TwGL, there is a pressing demand for a comprehensive survey reviewing current work from a unified view. In this survey, we provide a comprehensive overview of existing achievements in the TwGL field from four key dimensions, namely, robustness, explainability and privacy protection. For each part, we give a general categorization for existing work and review typical work for each category. To give further insights for TwGL research, we provide a unified view to inspect previous work and build the connections between them. We also point out some important open problems remaining to be solved in the future developments of TwGL.


\nosection{Overview}
In this paper, we puts together the state-of-the-art and most important works for TwGL. The organization of this paper is structured as follows: 
In Section \ref{sec:robustness}, we comprehensively review the recent advances of reliable DGL against three threats, including inherent noise, distribution shift, and adversarial attack. In Section \ref{sec:explainability}, we provide a detailed technique overview of the state-of-the-art GNN explainability methods. In Section \ref{sec:privacy}, we also discuss the research on the privacy protection of DGL models. 
For each section, we summarize several challenges and opportunities for future research. Finally, we conclude this paper by providing insights for further achievements of trustworthy graph learning in Section \ref{sec:conclusion}.

\nosection{Relation to other surveys}
To our best knowledge, little survey have been comprehensively conducted especially on the methods and applications of TwGL.
There are several surveys related to our paper, most of which focus on a single aspect of TwGL, e.g., reliability \cite{DBLP:journals/corr/abs-1812-10528,DBLP:journals/corr/abs-2003-05730,DBLP:journals/corr/abs-2003-00653,xu2021robustness,DBLP:journals/corr/abs-2103-03036,DBLP:journals/corr/abs-2112-06070} and explainability \cite{yuan2020explainability,DBLP:journals/corr/abs-2203-09258}.  
Comparatively to the aforementioned surveys, we present a systematic and comprehensive review of the current advances, trends, and applications in trustworthy graph learning, which would potentially benefit the research community in developing practical trustworthy models. 

\nosection{Basic definitions}
Oracle graph data $\mathcal{G} = (\mathbf{A}, \mathbf{X})$ with its adjacency matrix $\mathbf{A}$ and node features $\mathbf{X}$; $Y$ the labels of graphs/nodes; $\epsilon_a, \epsilon_x, \epsilon_y$ denote structure, attribute, and label noises respectively; $\mathcal{D}^{\text{train}}$ and $\mathcal{D}^{\text{test}}$ are the training and test dataset, respectively; $\mathcal{L}$ denotes the loss function and $\hat{\mathcal{G}}$ is the perturbed graph.

\section{Reliability}
\label{sec:robustness}

\subsection{Overview}

\begin{figure*}[h]
    \centering
    \includegraphics[width=\linewidth]{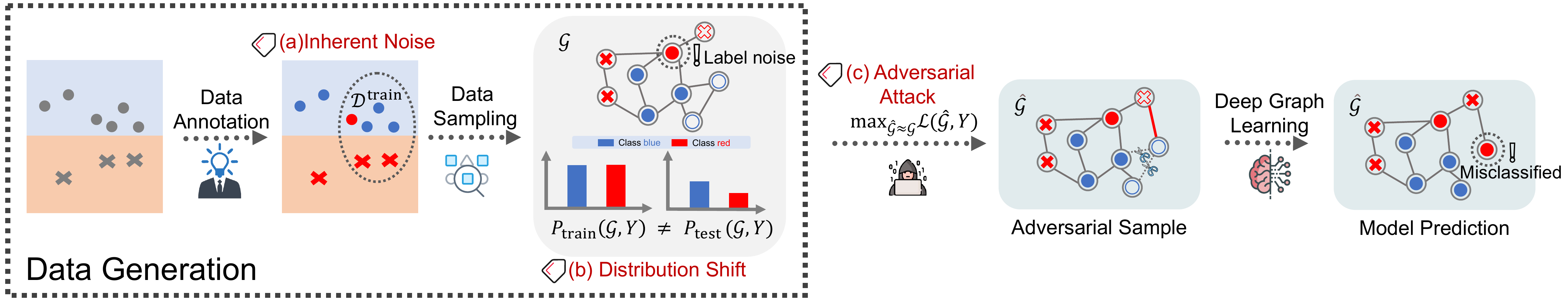}
    \caption{An illustrative example on deep graph learning against (a) inherent noise, (b) distribution shift, and (c) adversarial attack. From left to right: the inherent noise (i.e., label noise) is inevitably introduced during data annotation, where a red ``circle'' node is mislabeled with the ``cross'' class (marked with red color). Then, the sampling bias leads to the discrepancy between training (color-filled nodes) and testing datasets (color-unfilled nodes), thus introducing any possible kind of distribution shift.
    After the data generation process, adversarial attacks 
    are performed on the graph to mislead the model prediction (e.g., misclassification on a node).
}
    \label{fig:framework}
\end{figure*}

\begin{table*}[htp]
    \centering
    \caption{Summary of recent advances in reliable deep graph learning. 
    }
    \label{tab:reliable_gnn_overview}
    \small
    \begin{tabular}{ccc}
        \toprule
        Threats & Description & Selected References \\
        \midrule
        
        \multirow{3}{*}{Inherent Noise} &  \multirow{3}{*}{$\mathcal{D}^{\text{train}}=(\mathbf{A}+\epsilon_a, \mathbf{X}+\epsilon_x, Y+\epsilon_y)$} & \cite{zheng2020robust},\cite{luo2021learning},\cite{nt2019learning},\\ 
        & & \cite{patrini2017making},\cite{li2021unified},\cite{dai2021nrgnn}\\ 
        & & \cite{rong2019dropedge},  \cite{verma_graphmix_2020}
        \\
        \midrule
        \multirow{4}{*}{Distribution shift} & \multirow{4}{*}{$P_{\text{train}}(\mathcal{G}, Y) \neq P_{\text{test}}(\mathcal{G}, Y)$} & \cite{DBLP:conf/icml/BevilacquaZ021}, \cite{wu2022towards}, \cite{DBLP:journals/corr/abs-1908-05429}, \\ 
        & & 
        \cite{DBLP:journals/corr/abs-2106-11133}, \cite{kose2022fair}, \cite{DBLP:journals/corr/abs-2010-09891}, \\ 
        & & \cite{DBLP:conf/www/WuP0CZ20}, \cite{han2021reliable}, \cite{wang2021confident} \\ 
        \midrule
        \multirow{3}{*}{Adversarial Attack}  & \multirow{3}{*}{$\max_{\hat{\mathcal{G}} \approx \mathcal{G}} \mathcal{L}(\hat{\mathcal{G}}, Y) $}  & 
        \cite{Wu0TDLZ19}, \cite{EntezariADP20}, \cite{jin2020graph},\\
        & & \cite{DBLP:conf/icml/LiuJ0LLW0T21}, \cite{geisler2021robustness}, \cite{chen2021understanding},\\
        & & \cite{XuC0CWHL19}, \cite{DBLP:journals/tkde/FengHTC21}, \cite{DBLP:conf/wsdm/TangLSYMW20}\\
        \bottomrule
    \end{tabular}
\end{table*}
Recent few years have seen deep graph learning (DGL) based on graph neural networks (GNNs) making remarkable progress in a variety of important areas, ranging from business scenarios such as finance (e.g., fraud detection and credit modeling) \cite{fraud_gnn,credit_gnn, bian2020rumor,li2019semi},  e-commerce (e.g., recommendation system) \cite{fedgnn_rec,min2022masked}, drug discovery and advanced material discovery \cite{guo2021dockstream,molecularnet,rong2020self,ma2022cross,mao2021molecular}.
Despite the progress, applying various DGL algorithms to real-world applications faces a series of reliability threats. 
At a high level, we categorize these threats into three aspects, namely,  \emph{inherent noise}, \emph{distribution shift}, and \emph{adversarial attack}. 
Specifically, inherent noise refers to irreducible noises
in graph structures, node attributes, and node/graph labels. Distribution shift refers to the shift between training and testing distribution which includes both domain generalization and sub-population shift. 
Adversarial attack is a manipulative human action that aims to cause model misbehavior with carefully-designed patterns or perturbations on the original data
Typical examples include adversarial samples \cite{DBLP:conf/kdd/ZugnerAG18} and backdoor triggers \cite{DBLP:conf/uss/XiPJ021}.

Figure \ref{fig:framework} visualizes how different threats occur throughout a typical pipeline of deep graph learning. 
As a comparison, inherent noise or distribution shift typically happens in the data generation process due to sampling bias or environment noise without deliberate human design while adversarial attacks are intentionally designed by malicious attackers after the data generation phase.
Table \ref{tab:reliable_gnn_overview} further provides formal descriptions of each threat, and summarizes some typical related work.

\begin{figure*}[h]
    \centering
    \includegraphics[width=0.9\linewidth]{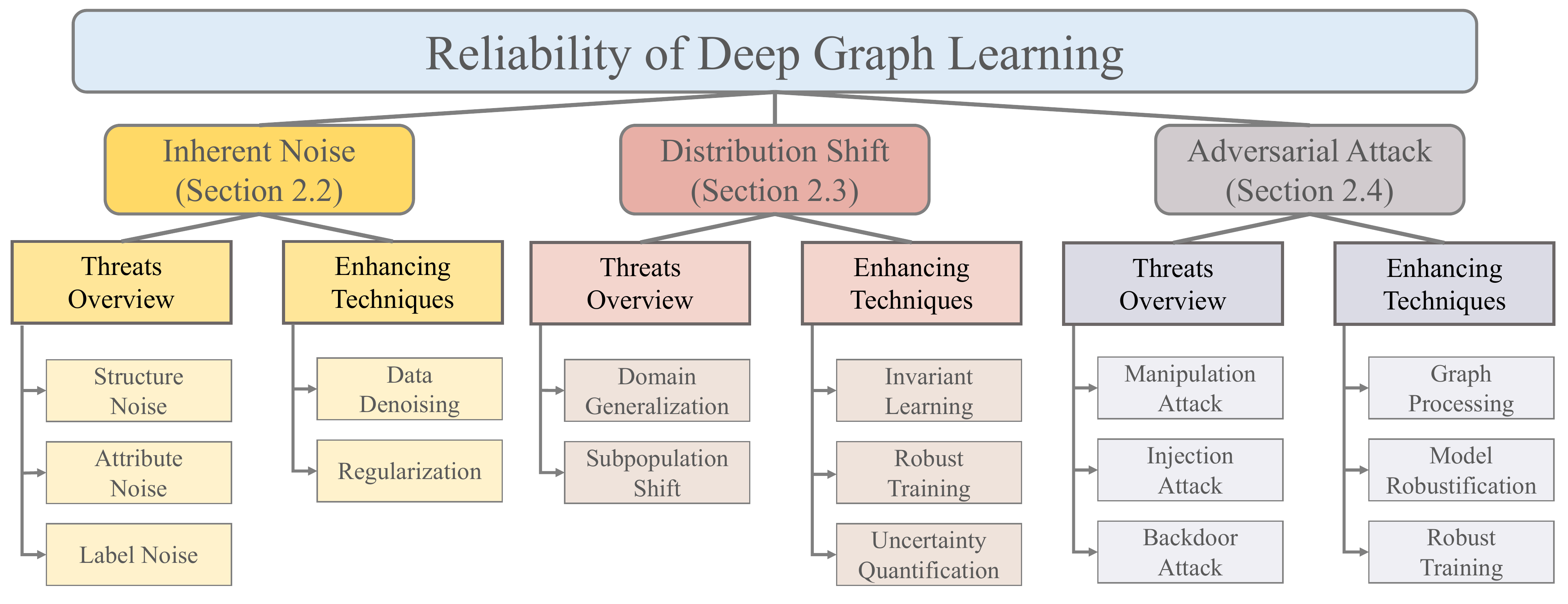}
    \caption{An overview of reliable deep graph learning.}
    \label{fig:overview}
\end{figure*}

Numerous work emerges to improve the reliability of DGL algorithms against the above threats from different perspectives such as optimization policy design and uncertainty quantification.
In this section, we explore recent advancements in this research direction. We provide the overview of reliable deep graph learning in Figure \ref{fig:overview}.

\subsection{Reliability against Inherent Noise}

The inherent noise refers to the noise that inherently exists in graph data.
It is unavoidable that real-world graph data would include some inherent noises, which might stem from error-prone data measurement or collection, information loss during data preprocessing, sub-optimal graph structure w.r.t. downstream tasks, and so on. This section first discusses how inherent noises can affect DGL algorithms, and then summarizes the recent typical techniques that can explicitly or implicitly alleviate the effects caused by inherent noises occurring in the pipeline of DGL.

\subsubsection{Inherent Noise}
Inherent noises in graph data may exist in graph structure, node attributes, and node/graph labels, which draws out three types of inherent noises, i.e., structure noise, attribute noise, and label noise. They are detailed as follows.

\nosection{Structure noise} The inherent structure noise may inevitably be introduced due to error-prone data measurement or collection \cite{chen2020iterative,wang2019learning,xu2020robust}, e.g., protein-protein interaction networks
\cite{FIONDA2019915}. And it could also come from sub-optimal graph construction with task-irrelevant edges \cite{chen2020iterative,luo2021learning}, as the underlying motivation of establishing edges might not be always relevant to a specific downstream task \cite{zheng2020robust}. Since most existing DGL algorithms rely on the input structure to propagate information, the structure noise presents negative effects on the propagating process and consequently the final model performance.

\nosection{Attribute noise} There are two kinds of inherent attribute noise. On the one hand, the raw attributes of nodes may be incomplete or incorrect \cite{wang2019learning}. For example, users may intentionally provide fake profiles due to privacy concerns in social networks. On the other hand, another subtle source is information loss when transforming raw node attributes into node embeddings, e.g., using the word embedding technique to encode textual data of nodes. During the training phase of message-passing based GNNs, the attribute noise in each node can be passed to its neighbors, which would gradually affect the final node/graph embeddings.

\nosection{Label noise} The erroneous annotations in node or graph labels lead to the inherent label noise. Similarly to image classification, unavoidable erroneous annotations can occur in node classification. But comparing to images, it might be harder to annotate labels for nodes due to node dependency in graphs, which results in more noises or label sparsity issues \cite{li2021unified,nt2019learning}. In graph classification, annotating labels for graphs becomes more expensive because of requiring more domain expertise. For instance, to annotate a drug-related property of a molecular graph, it needs to conduct experiments for measurement or extract the property from literature, which are both likely to introduce graph label noise \cite{ji2022drugood}. It is worth noting that, the previous study has shown that deep learning models tend to overfit label noise, which is also likely to happen in DGL models \cite{li2021unified,dai2021nrgnn}. 

\subsubsection{Enhancing Techniques}
A number of methods have been presented for enhancing the robustness of DGL algorithms against above inherent noises. Although some of these methods are not originally designed for mitigating inherent noises, their behind ideas are useful for building robustness enhancing techniques, and are thus also covered in this survey. In this subsection, we review these methods from data and regularization perspectives.

\nosection{Data denoising}
From the data perspective, data denoising is a straightforward way for reducing the noise effects in DGL algorithms. 
For \emph{structure noise}, one natural idea is to assign learnable weights for each node when perform node information aggregation in GNNs, so that the weight can hopefully reflect the tasks-desired graph structures. One of well-known techniques is incorporating self-attention modules into the original GNNs \cite{chen2020iterative}. As a more explicit way to denoise graph data, \cite{zheng2020robust} and \cite{luo2021learning} propose to prune task-irrelevant edges using a multi-layer neural network to draw a subgraph from a learned distribution. The reparameterization trick \cite{DBLP:conf/iclr/JangGP17} is applied to make this process differentiable, which can thus be jointly trained with its subsequent GNNs for a specific task. 
Regarding \emph{label noise}, \cite{nt2019learning} first observe label noise can lead to significant performance drops in the DGL setting. Motivated by prior methods for non-graph machine learning, this work directly applies the backward loss label correction \cite{patrini2017making} to train GNNs without considering graph structure. More recently, some work incorporates the structure information into the design. \cite{li2021unified} conduct sample reweighting and label correction by using random walks over graph structure for label aggregation to estimate node-level class probability distributions. \cite{dai2021nrgnn} propose to generate accurate pseudo labels, and assign high-quality edges between unlabeled nodes and (pseudo) labelled nodes to reduce label noise. 
Despite the existence of \emph{attribute noise}, to our best knowledge, there is currently no  methods specially designed for denoising attribute noise, which can be a promising future direction.
In addition, it is interesting and probably more useful to simultaneously diminish multiple noises from structure, attributes, and labels, since denoising one type of noises often require other type(s) information (desirably without noises) for assistance.

\nosection{Regularization}
From the regularization perspective, previous studies attempt to reduce overfitting to inherent noises by decreasing the complexity of the hypothesis space implicitly or explicitly. 
On the one hand, various regularization techniques specially for GNNs are presented for reducing overfitting of DGL algorithms. The core idea behind these techniques
is to randomly drop edges \cite{rong2019dropedge}, nodes \cite{hamilton2017inductive}, or hidden representations of GNNs \cite{chen2018fastgcn}. \cite{hasanzadeh2020bayesian} take a further step to unify these approaches into a Bayesian graph learning framework. Besides, an advanced data augmentation strategy namely Mixup is recently applied to DGL and is proved to be effective for several graph-related tasks \cite{verma_graphmix_2020,wang2021mixup,wu2021graphmixup}.
On the other hand, there are some studies trying to explicitly impose regularization to the hypothesis space, i.e., building GNN models with some predefined constrains or inductive bias, which share the similar idea with approaches demonstrated in Section \ref{subsec:adv_enh}. 
$p$-Laplacian based GNNs \cite{fu2021p} introduces a discrete $p$-Laplacian regularization framework to derive a new message passing scheme and impose GNNs to be effective for both homophilic and heterophilic graphs as well as improved the robustness on graphs with noisy edges.
Moreover, from the Bayesian perspective, prior ensemble approach for DGL can also be seen as an implicit regularization, which independently trains multiple graph learning models with different input data transformations, then aggregates outputs from multiple models as the final output \cite{papp2021dropgnn,hou2021robust}.

\subsection{Reliability against Distribution Shift}

Distribution shift in machine learning occurs when the training data and test data are generated by the different underlying distributions, which exists in a lot of real-world applications.
Distribution shift on classic data format (e.g. vision or texts) have been comprehensively investigated in some recent surveys (e.g., \cite{DBLP:conf/ijcai/0001LLOQ21,zhou2021domain}).
However, there are inadequate discussions on graph data.
Here we provide a review of distribution shift literature with a focus on graph-structured data. This section first provides a categorization for typical distribution shift on graph data then introduce recent work for improving the reliability of DGL methods against distribution shift.

\subsubsection{Distribution shift on graphs}
Motivated by prior work focusing on distribution shift in general machine learning, we categorize distribution shift on graph-structured data into two types, \emph{domain generalization} and \emph{sub-population shift}. Domain generalization refers to the training and testing distributions consist of distinct domains. Some typical examples include covariate shift and open-set recognition. While sub-population shift refers to training and testing distributions consist of the same group of domains but differ in the frequencies of each domain.

\nosection{Domain generalization}
One typical example of domain generalization on graph-related tasks is covariate shift, which assumes that the conditioned label distribution is the same for both training and test domains but differs in the data marginal distribution~\cite{DBLP:conf/icml/BevilacquaZ021}. For examples, in drug discovery, the scaffolds of drug molecules often differ at inference and in social networks and financial networks, the graph structures may significantly change with time~\cite{wu2022towards}. 

\nosection{Sub-population shift}
Sub-population shift on graphs raises when the frequencies of data/label sub-populations change~\cite{kose2022fair}, which widely exists in many graph learning tasks, such as algorithmic fairness  and label shift.
Specifically, fairness issues of DGL could be caused by  societal bias contained in graph data~\cite{kose2022fair}. 
Label shift refers to the cases where the marginal label distribution changes for two domains but the conditional distributions of the input given label stay the same across domains~\cite{DBLP:journals/corr/abs-2106-11133}. In addition, class-imbalance problem on graphs is a specific form of label shift~\cite{DBLP:journals/corr/abs-2106-11133}. For instance, the label distribution is uniform for the testing distribution but is not for the training distribution.

\subsubsection{Enhancing techniques}
\label{sec:dshift_en}
Recently, there have been some methods proposed to tackle the challenges raised by distribution shift on graphs, which could mainly be classified into three categories: invariant representation learning, robust training, and uncertainty quantification.  

\nosection{Invariant learning}
Invariant learning aims to learn invariant graph representations across different domains. The idea of invariant representation learning proposed recently has been adapted in some DGL models. 
\cite{DBLP:conf/icml/BevilacquaZ021} use a causal model to learn approximately invariant graph representations that well extrapolate between the training and testing domains.
\cite{wu2022towards} inherit the spirit of invariant risk minimization~\cite{DBLP:journals/corr/abs-1907-02893} to develop an explore-to-extrapolate risk minimization framework that facilitates GNNs to leverage invariant graph features for node-level prediction. 
\cite{DBLP:journals/corr/abs-1908-05429} design an adversarial domain classifier to learn the domain-invariant representation for network alignment, which is optimized by simultaneously minimizing its loss and maximizing a posterior probability distribution of the observed anchors.
\cite{DBLP:journals/corr/abs-2106-07482} present a disentanglement-based unsupervised domain adaptation method which applies variational graph auto-encoders to recover three types of defined latent variables (semantic, domain, and random latent variables). 


\nosection{Robust training}
Graph robust training proposes to enhance the model robustness against distribution shift either by data augmentation or modifying the training framework. On the one hand, some methods generalize advanced data augmentation techniques for general data format (e.g., mixup) to graph-structured data. \cite{DBLP:journals/corr/abs-2106-11133} present a mixup-based framework for improving class-imbalanced node classification on graphs, which performs feature mixup on a constructed semantic relation space and edge mixup. \cite{kose2022fair} propose a fairness-aware data augmentation framework on node features and graph structure to reduce the intrinsic bias of the obtained node representation by GNNs. On the other hand, some methods propose to integrate adversarial training techniques in DGL models. \cite{DBLP:journals/corr/abs-2010-09891} present a method that iteratively augments node features with adversarial perturbations during training and helps the model generalize to out-of-distribution (OOD) samples by making it invariant to small fluctuations in input data. 
\cite{DBLP:conf/www/WuP0CZ20} utilize the attention mechanism to integrate global and local consistency and the gradient reversal layer~\cite{DBLP:conf/icml/GaninL15} to learn cross-domain node embeddings. 

\nosection{Uncertainty quantification}
Apart from the above two directions that aim to improve model robustness, uncertainty quantification can be seen as a complementary way to enhance the reliability of DGL algorithms, since the estimated uncertainty can be used for rejecting unreliable decisions with high model predictive uncertainties. Here we introduce some recent work of uncertainty quantification for DGL algorithms. 
A natural uncertainty measure can be the prediction confidence, i.e., the maximum value of the Softmax output. However, recent work observes that GNNs with Softmax prediction layer are typically under-confident, thus the confidence cannot precisely reflect the predictive uncertainty \cite{wang2021confident}. There are two ways to solve this problem. 
First, recent work introduces probabilistic blocks into original GNNs for modeling the posterior weight distribution, which can provide more accurate uncertainty estimation than deterministic GNN architectures. For example, \cite{han2021reliable} propose to replace the Softmax decision layer with a Gaussian process block, which provides accurate uncertainty estimations. Unlike this work, Bayesian GNNs \cite{hasanzadeh2020bayesian} aggressively transforms whole GNN layers into Bayesian counterparts. Concretely, it treats both model weights and sampling process as probabilistic distributions and adopts variational inference to estimate the parameters of these distributions.
Second, a more direct way is to perform confidence calibration in a post-hoc fashion without modifying the GNN architectures. One typical calibration method is temperature scaling. However, it is originally designed for DNNs and is proved to have poor performance in the DGL setting. \cite{wang2021confident} modify temperature scaling for GNNs by using an additional GNNs to predict a unique temperature for each node. Since the temperatures are produced by considering both node features and graph topology, this method achieves better calibration performance compared to the original method.

\subsection{Reliability against Adversarial Attack}
Adversarial attacks aim to cause a model to make mistakes with carefully-crafted unnoticeable perturbations (adversarial samples) or predefined patterns (backdoor triggers). Although promising results have been achieved, recent studies have shown that GNNs are vulnerable to adversarial attacks, posing significant security risks to several application domains \cite{DBLP:journals/corr/abs-1812-10528}.
Therefore, the adversarial reliability of GNNs is highly desired for many real-world systems, especially in security-critical fields. In this section, we provide an overview of adversarial attacks to GNNs and subsequently review recent works that mitigate such threats.

\subsubsection{Threats overview}
Literature has categorized adversarial attacks into several typical dimensions \cite{DBLP:journals/corr/abs-1812-10528,DBLP:journals/corr/abs-2003-05730}. Specifically, adversarial attacks can be performed in the training phase (poisoning attack) and the inference phase (evasion attack), to mislead the prediction of the model on specific important instances such as nodes (targeted attack), or degrade the overall performance of the model (non-targeted attack). 
As shown in Table \ref{tab:reliable_gnn_overview}, the goal of adversarial attacks is to maximize the loss of DGL models with imperceptible perturbations on the original graph.  
Based on the way employed by attackers to perturb the graph data or learning model, this survey reviews prior work from three dimensions: \emph{manipulation attacks}, \emph{injection attacks} and \emph{backdoor attacks}. Specifically, manipulation and injection attacks can be performed in the inference phase while the backdoor attacks always happen in the training phase.



\nosection{Manipulation attacks}
In manipulation attacks, attackers generate adversarial samples by modifying either the graph structure or node attributes. For instance, attackers can add, remove or rewire an edge in the graph to generate legitimate perturbations. 
As a pioneering work, \cite{DBLP:conf/kdd/ZugnerAG18} craft adversarial samples by manipulating both edges and attributes of the graph with a greedy search algorithm. Followed by this work, the gradient-based approach becomes a prominent way to craft adversarial samples. By exploiting the gradient information of the victim model or a locally trained surrogate model, attackers can easily approximate the worst-case perturbations to perform attacks
\cite{li2021adversarial,XuC0CWHL19,Wu0TDLZ19}. While current research heavily relies on supervised signals such as labels to guide the attacks and are targeted at certain downstream tasks, \cite{zhang2022unsupervised} propose an unsupervised adversarial attack where gradients are calculated based on graph contrastive loss. In addition to gradient information, \cite{DBLP:conf/aaai/ChangRXHZC0H20} 
proposed to approximate the graph spectrum and propose a generalized adversarial attacker \textbf{GF-Attack} to perform attacks in a black-box fashion.  \cite{chang2022adversarial} further prove that GF-Attack can perform the effective attack without knowing the number of layers / embedding window size of the original model. 


\nosection{Injection attacks}
The manipulation attack requires the attacker to have a high privilege to modify the original graph, which is impractical for many real-world scenarios. Alternatively, injection attacks has recently emerged as a more practical way that injects a few malicious nodes into the graph. The goal of the injection attack is to spread malicious information to the proper nodes along with the graph structure by several injected nodes. In the poisoning attack setting, \cite{DBLP:conf/www/SunWTHH20} first study the node injection attack and propose a reinforcement learning based framework to poison the graph. \cite{DBLP:journals/datamine/WangLSLYZ20} further derive an approximate closed-from solution to linearize the attack model and inject new vicious nodes efficiently. \cite{DBLP:conf/ijcai/ZhangZGMSL019} extends the poisoning attack from homogeneous graphs to heterogeneous graphs.  In the evasion attack setting, \cite{DBLP:conf/kdd/ZouZDGKLT21}
consider the attributes and structure of injected nodes simultaneously to achieves better attack performance. However, a recent work \cite{chen2022understanding} shows that the success of injection attacks is built upon the severe damage to the homophily of the original graph. Therefore, the authors present to optimize a harmonious adversarial objective to preserve the homophily of graphs.  

\nosection{Backdoor attacks}
In contrast to prior two attacks, backdoor attacks aim to poison the learned model by injecting \emph{backdoor triggers} into the graph at the training stage. As a consequence, a backdoored model would produce attacker-desired behaviors on trigger-embedded inputs (e.g., misclassify a node as an attacker-chosen target label) while performing normally on other benign inputs. Typically, a backdoor trigger can be a node or a (sub)graph designed by attackers.  \cite{DBLP:conf/uss/XiPJ021} and \cite{DBLP:conf/sacmat/ZhangJWG21} propose to use subgraphs as trigger patterns to launch backdoor attacks. \cite{DBLP:conf/wisec/XuXP21} select the optimal trigger for GNNs based on explainability approaches. Backdoor attacks are a relatively unexplored threat in the literature. However, they are more realistic and dangerous in many security-critical domains as the backdoor trigger is hard to notice even by human beings.

\subsubsection{Enhancing techniques}
\label{subsec:adv_enh}
Efforts have been made to mitigate adversarial attacks in different ways.
In this subsection, we review recent robustness enhancing techniques of DGL against adversarial attacks from data, model and optimization perspectives.

\nosection{Graph processing}
From data perspective, a natural idea is to process the training/testing graph to remove adversarial noises thus mitigating its negative effects.  
Currently, enhancing approaches in this direction is mainly supported by empirical observations on specific adversarial attacks \cite{Wu0TDLZ19,EntezariADP20,jin2020graph,li2022guard}. For example, there is a tendency of adversarial attacks to add edges between nodes with different labels and low attribute similarity. In this regard, 
\cite{Wu0TDLZ19} prune the perturbed edges based on Jaccard similarity of node attributes with the assumption that adversarially perturbed nodes have low similarity to some of their neighbors.
Another work \cite{EntezariADP20} observes that the adjacent matrix of adversarially perturbed graphs are always with high rank. Based on this observation, they employ SVD decomposition to form a low-rank approximation for the adjacent matrix thus can reduce the effects of adversarial attacks to a certain degree. Further, the clean graph structure can be learned simultaneously by preserving graph properties of sparsity, low rank, and feature smoothness during training \cite{jin2020graph}. 
Graph processing based methods are cheap to implement while significantly improving the adversarial robustness of GNNs. However, empirical observation based on specific attacks makes such methods difficult to resist unseen attacks. 
 In addition, there is a particular trade-off between performance and robustness since they often hold the assumption that data has already been perturbed.

\nosection{Model robustification}
Refining the model to prepare itself against potential adversarial threats is a prominent enhancing technique and we term it as model robustification. Specifically, the robustification of GNNs can be achieved by improving the \emph{model architecture} or \emph{aggregation scheme}.  There are several efforts that aim to improve the architecture by employing different regularizations or constraints on the model itself, such as 1-Lipschitz constraint \cite{DBLP:conf/icml/ZhaoZZWJZJD021}, $\ell_1$-based graph smoothing \cite{DBLP:conf/icml/LiuJ0LLW0T21} and adaptive residual \cite{liu2021graph}. As recently shown in \cite{geisler2021robustness,chen2021understanding}, the way that GNNs aggregate neighborhood information for representation learning makes them vulnerable to adversarial attacks. To address this issue, they derive a robust median function instead of a mean function to improve the aggregation scheme. Overall, a robustified model is resistant to adversarial attacks without compromising on performance in benign situations.

\nosection{Robust training}
Another enhancing technique successfully applied to the GNN model is based on the robust training paradigms. Adversarial training is a widely used practical solution to resist adversarial attacks, which builds models on a training set augmented with handcrafted adversarial samples. Essentially, the adversarial samples can be crafted via specific perturbations on the graph structure \cite{XuC0CWHL19}, node attributes \cite{DBLP:journals/tkde/FengHTC21} and even hidden representations \cite{jin2019latent}. Although adversarial training can improve the generalization capability and robustness of a model against unseen attacks, there is a risk of overfitting to adversarial samples.
In addition, adversarial training is only able to resist evasion attacks. To this end,  \cite{DBLP:conf/wsdm/TangLSYMW20} propose to enhance the model robustness against poisoning attacks through transferring knowledge from similar graph domains. However, it requires a large number of clean graphs from other domains to successfully train the model.  \cite{chang2021not} provides a detailed theoretical analysis of the robustness of the low-frequency components in the graph convolution filter and reveals that not all low-frequency components are robust to adversarial attacks. In this vein, \cite{chang2021not} proposes a new robust co-training paradigm: \textbf{GCN-LFR}, which can transfer the robustness from the low-frequency components within eligible robust interval with a auxiliary regularization net. 


\subsection{Overall Discussion}
\label{sec:reliable_discusion}
Given the above comprehensive summary of recent advances in reliable DGL research, we further provide overall discussions among the above topics including their relations, differences, and applications.

\nosection{Difference to general reliable machine learning}
Tremendous efforts have been made to improve the reliability of deep learning on non-graph data such as images and texts. In contrast to these methods, improving the reliability of GNNs on graph data poses unique challenges.
For the message-passing-based models used by most DGL algorithms,
the adversarial perturbation, inherent noise and distribution shift of one node can be transmitted to its neighbors and further hinder the model performance. Therefore, previous general reliable machine learning algorithms need to be modified by considering the relationship between different nodes.

\nosection{A unified view}
The uncertainty modeling framework allows us to examine the three types of threats in a unified manner.
There are two typical types of uncertainties, \emph{aleatoric} and \emph{epistemic} uncertainties.
Specifically, we can treat inherent noise as the main source of aleatoric uncertainty, since these noises are irreducible even given access to infinite samples. 
In addition, we can treat adversarial attack and distribution shift as two sources of epistemic uncertainty, as we can reduce these uncertainties by introducing data samples on different domains/sub-populations through advancing data augmentation.
For example, to combat with adversarial noises, robust training \cite{XuC0CWHL19} involves adversarial samples into the training process to enhance the adversarial robustness of GNN models.
Under this unified view, we can leverage prior uncertainty estimation methods to detect adversarial samples (adversarial attack), out-of-distribution samples (distribution shift), and outliers (inherent noises), which provides further information for enhancing model's reliability.

\nosection{Difference among above threats}
In general, the above three types of threats can all be seen as the ``mismatch'' between training and testing samples. Here, we highlight subtle differences among them. The inherent noise or distribution shift happens in the training data generation process due to sampling bias and environment noise without deliberate human design, while adversarial attacks are deliberately designed by malicious attackers after the training/testing sample generation phase. Furthermore, the inherent noise is typically irreducible, while the other two types of threats can be alleviated by sampling more data with expertise. Despite the differences, some general techniques can be used for preventing the three types of threats. For example, one can inject probabilistic decision module into GNNs to provide predictions and its uncertainty estimation \cite{han2021reliable,hasanzadeh2020bayesian}. 
Uncertainty estimation can be further enhanced by introducing OOD and adversarial samples into the training phase. 
Thus, the estimation can be used to detect the above threats, improving the decision reliability of various DGL algorithms.

\subsection{Future  Directions}

This part gives an overview of recent advances for improving the reliability of DGL algorithms in terms of 
the following three aspects: adversarial attack, inherent noise, and distribution shift.
For each type of threat, we provide a systematical view to inspect previous robustness-enhancing techniques for DGL. 
Our survey can help researchers to better understand the relationship between different threats and to choose appropriate techniques for their purposes.
Despite the above progress, there still exist several interesting future directions as summarized below.

\nosection{Theoretical framework}
 Despite algorithmic advances for reliable DGL, there is still a lack of theoretical frameworks to formally analyze the effectiveness of these methods. For example, how to analyze out-of-distribution generalization bound in the DGL setting remains an open problem.
 
 \nosection{Unified solution}
Section \ref{sec:reliable_discusion} shows relations among three threats. In a real-world setting, these threats may happen simultaneously. Therefore a unified solution to mitigate the effects caused by these threats is desired.

\nosection{Connection to learning stability}
As pointed out by prior work for general ML, there is a strong connection between robustness and learning stability. Thus, from the optimization perspective, how to build robust learning algorithm for DGL is also an interesting direction. Prior work \cite{wu2022towards} mentioned in Section \ref{sec:dshift_en} can be seen as an initial attempt, which applies invariant risk minimization to DGL settings to learn a stable graph representation. 

\nosection{Scalability and adversarial robustness}
Existing studies of the reliability to adversarial attacks mainly focus on relatively small graphs 
How to transfer these methods to real-world large-scale graph remains unexplored.

\nosection{Fairness and adversarial robustness}
 Existing work prefers to utilize standard performance metrics, such as accuracy, to measure the robustness of a DGL model, 
 which is, however, apparently insufficient for evaluating the overall reliability performance. 
 A DGL algorithm with high accuracy may not be fair to different attribute groups, which results in severe disparities of accuracy and robustness between different groups of data. 
This calls for future work to explore fair and robust DGL algorithms and develop principled evaluation metrics beyond accuracy.

 \nosection{Counterfactual explanations and adversarial samples}
One problem of DGL algorithms is the lack of interpretability. 
To address this issue, counterfactual explanations are proposed as a powerful means for understanding how decisions made by algorithms. While prior research in vision tasks \cite{DBLP:journals/corr/abs-2106-09992} has shown that counterfactual explanations and adversarial samples are strongly related approaches with many similarities, there is currently little work on systematically exploring their connections in DGL. 
 
 \nosection{Reliability benchmarks}
 Along with the fast development of reliable DGL algorithms, real-world benchmarks are desired for the research community.
 Some early benchmarks for specific settings have been established, e.g., \cite{ji2022drugood} present an OOD dataset curator and benchmark for AI-aided drug discovery designed specifically for the distribution shift problem with data noise. It is promising to build a general benchmark platform to cover more reliability aspects mentioned in this survey.

\section{Explainability}
\label{sec:explainability}

The emergence of Graph Neural Networks (GNNs) \cite{kipf2016semi} has revolutionized deep learning on graph-structured data. GNNs have achieved superior performances across many fields such as social network analysis \cite{rong2019dropedge,bian2020rumor}, biochemistry \cite{yu2022structure}, computer vision \cite{li2019deepgcns} and finance \cite{cheng2022financial}. 
Despite their great success, GNNs are generally treated as black-box since their decisions   are less understood \cite{ying2019gnnexplainer, luo2020parameterized}, leading to the increasing concerns about the explainability of GNNs. 
It is hard to fully trust the GNN-based models without interpretation to their predictions, and thus restrict their applications in high-stake scenarios such as clinical diagnose \cite{jaume2021quantifying, yu2021towards}, legal domains \cite{yang2021legalgnn} and so on. 
Hence, it is imperative to develop the explanation techniques for the improved transparency of GNNs.
\begin{figure*}[t]
\begin{center}
\centerline{\includegraphics[width=1.0\columnwidth]{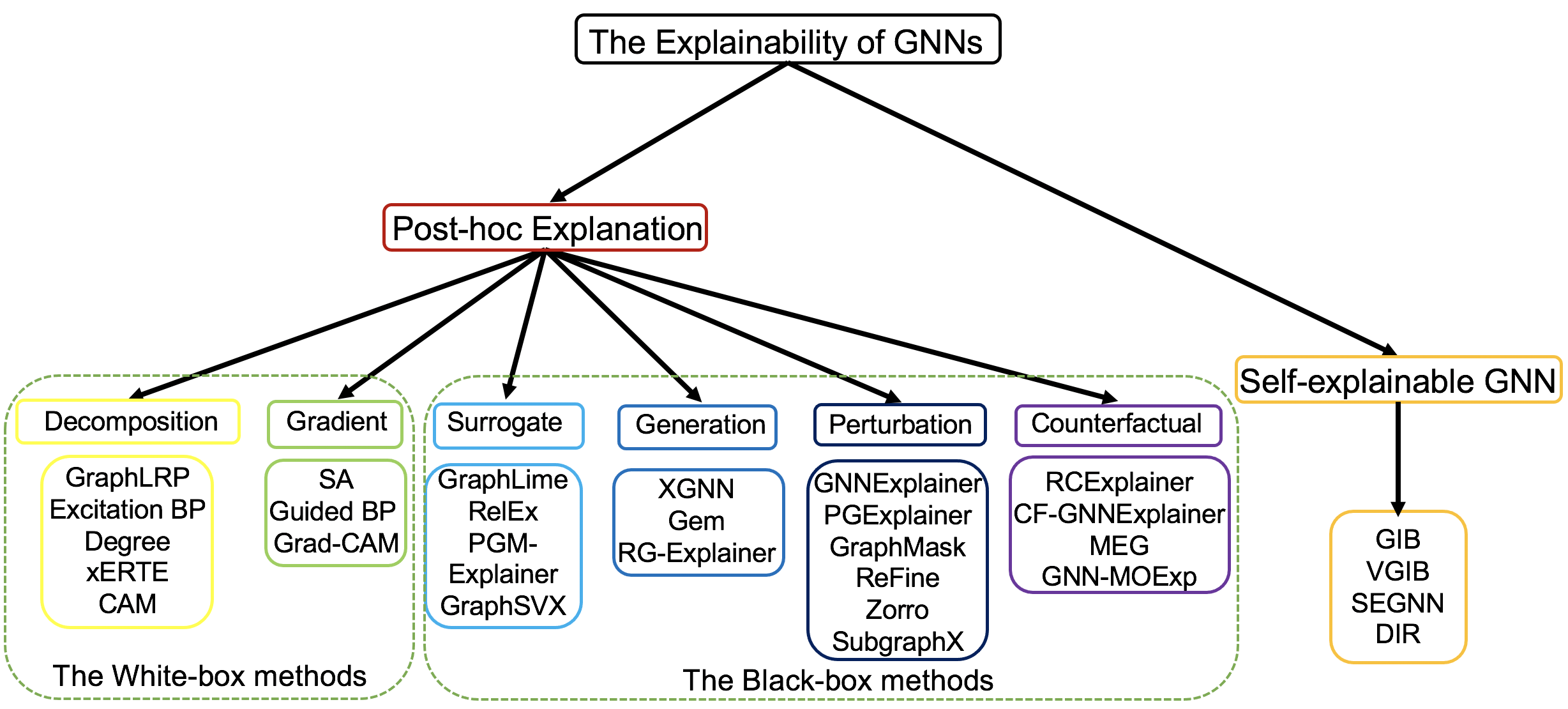}}
\end{center}
\caption{An overview of the explainability of graph neural networks.}
\label{taxonomy}
\end{figure*}
Recently, several methods are proposed to uncover the  knowledge which drives GNN's prediction on different aspects.
Generally, they highlight the important patterns of the input graphs such as nodes \cite{huang2020graphlime, pope2019explainability}, edges \cite{ying2019gnnexplainer, luo2020parameterized} and sub-graphs \cite{yu2020graph,yuan2021explainability} which are crucial for the model predictions.
In Figure~\ref{taxonomy}, we provide an overview of the explainability of GNNs.
Those explanation methods are categorized into two categories: post-hoc methods and self-explainable methods,
according to whether the explanation method is architecturally built-in within the graph model.

Given a pre-trained graph model, the post-hoc methods focus on discovering the crucial substructures that most influence the prediction. 
Based on the knowledge of the explanation methods, we further categorize the post-hoc methods into two sub-categories as the white-box methods and the black-box methods.
The white-box methods have to access the model parameters or gradients while the black-box methods assume no knowledge of GNNs' parameters or gradients but require the input and output of the graph model. 

Apart from the post-hoc explanation methods, the self-explainable methods have attracted increasing attentions in the recent literature. 
Compared with the post-hoc explanation methods, the self-explainable methods enjoy computational and time efficiency by making predictions and producing the corresponding explanations,  simultaneously. However, it also requires domain knowledge to design the built-in module for the intrinsic explanations.

\subsection{Post-hoc White-box Methods}
Based on the different knowledge obtained from the pre-trained graph model, the post-hoc white-box methods consists of two types of methods: 1) the decomposition-based methods which decompose the prediction into the contributions of different input substructures; and 2) the gradient-based methods, which generally employ the output-gradient or logits-gradient w.r.t the input graph to identify the importance of the input portions.

\subsubsection{Decomposition-based Methods}

\textbf{GraphLRP \cite{schwarzenberg2019layerwise}} extends the layer-wise relevance propagation (LRP) \cite{bach2015pixel} scheme to compute the node-level importance scores of the final prediction in terms of the weights and the hidden representations of the GNN layers. It tracks the information flow of the message-passing procedure in the GNNs, and then propagates the relevance of the intermediate neurons back to the input domain. GraphLRP first computes the contributions of the neurons in the fully-connected layer to the final prediction. Then, the neural contributions are layer-wisely unrolled to the neighborhoods of the central nodes. Althoug GraphLRP enjoys computational efficiency and trustful explanation, it has to assume that the model parameters are accessible, which limits the applications of this method in privacy protection scenarios. 

\textbf{Excitation BP \cite{pope2019explainability}} models the layer-wise relevance as the probability of each intermediate neuron. Then, the probability of the input features are calculated w.r.t the Bayesian rule in a layer-by-layer manner. The conditional probability is defined as the multiplication of the weights and the activation of the neurons. Similar to LRP, the negative activation are treated as zero. Since Excitation BP implements LRP from the Bayesian perspective, it shares similar advantages and limitations as GraphLRP.

\textbf{DEGREE \cite{DEGREE}} is a two-stage method which first decomposes the contributions of the node groups, and then generates subgraph-level explanation via the aggregation algorithm. The explanatory subgraph captures the topological information so that it is more human-intelligible than the node-level explanation. 
Instead of tracing the whole information flows of the intermediate neurons, DEGREE explicitly disentangles the group of node representations into two portions. By decomposing different scheme for the GNN layers, activation functions, and the pooling operations, the mostly contributing portion of the node group to the prediction is selected as the explanation portion, while the rest part of decomposition is dropped as the background portion. 
As there are exponentially many node groups in graph-structured data, it is intractable to compute the contributions of all the node groups. Hence, for a given subgraph, it computes the relevant contribution of its adjacent nodes. Then, the node with the highest contribution will be merged into the current subgraph. The above aggregation steps are repeated until it reaches a predefined budget.

\textbf{xERTE \cite{han2020explainable}} iteratively samples the relevant edges and nodes that are important to forecast future links in the temporal knowledge graph. At each time stamp, it computes the attention scores of the selected edges and propagate those attention scores for the adjacent nodes. Thus, the contribution on the future link of the query is decomposed into the relevance of the edges and nodes, which further refine the representation of the query for better predicted performance.
Unlike prior decomposition-based methods which produce post-hoc explanations, xERTE provides human-intelligible interpretation to the prediction with the improved performance. 

\textbf{CAM \cite{pope2019explainability}} identifies the important node features by decomposing the contributions of the node features in the last graph convolutional layer instead of the input domain. It requires a GNN model that contains a global average pooling (GAP) layer to obtain the final embedding, and a fully connected (FC) layer with a softmax outputs to the predicted results.
To interpret the prediction of the GNN model, CAM first computes the contributions of different final feature mappings from their corresponding weights in the FC layer in terms of different predicting classes. 
Then, the node importance is obtained by weighted summation of these contributions.
Notice that CAM only applies to the GNN model with GAP layer, it can be hardly generalized to other GNN models with different architectures. Moreover, CAM is only able to explain the output of GNN models for graph-level tasks.

\subsubsection{Gradient-based Methods}

\textbf{SA \cite{baldassarre2019explainability}} studies the sensitivity of a differentiable graph model via the square norm of the gradients w.r.t the input features, such as node features and edge features. It assumes that the feature with the higher gradient norm has higher influence on the prediction. Thus, the value of gradient norm reflects the importance of the input features. Although SA is a simple and effective method for interpretation, it suffers from a saturation problem \cite{shrikumar2017learning}. When the model output changes slightly due to an input perturbation, the gradient norm will hardly show the importance of the input features. Moreover, SA is unable to exploit the topological information of the graph-structured data via the gradient norms. Hence, it only generates feature-level or node-level explanation instead of subgraph-level explanation. However, subgraph-level explanation is more intuitive and human-understandable.

\textbf{Guided-BP \cite{baldassarre2019explainability}} further advances SA by clipping the negative gradients w.r.t the input features. The produced explanation is more concentrated on the positive contribution of the input feature and ignores the negative portions. Similar to SA, Guided-BP also suffers from the saturation problem.

\textbf{Grad-CAM \cite{pope2019explainability}} shares similar idea with CAM \cite{pope2019explainability} and only interpret the GNN models for graph-level tasks.
It employs the gradients instead of weights between the GAP layer and the FC layer to sum up different feature contributions. Then, these contributions are averaged to obtain the node importance for explanation. Different from CAM, Grad-CAM can be generalized to the GNN architecture without the GAP layer. But similar to CAM, Grad-CAM is unable to explain the model output on node-level tasks.

\subsection{Post-hoc Black-box Methods}
Unlike the white-box methods, the post-hoc black-box methods only deal with the input and of the graph model instead of the model parameters or the model gradients. There are 4 kinds of methods involved in the post-hoc black-box methods. 1) The surrogate-based methods leverage a simple and explainable model to fit the output space of the complex graph model. 2) Meanwhile, the generation-based methods either use generative models to synthesize the crucial patterns customized for an input, or generate key structures to globally explain the behavior of model predictions. 3) Moreover, the perturbation-based methods generally remove the unimportant edges and nodes so that the final prediction remains unchanged under such perturbations. 4) In the contrast, the counterfactual-based methods identify the minimal substructure of the input which would change the original prediction if removed. 

\subsubsection{Surrogate Methods}

\textbf{GraphLIME \cite{huang2020graphlime}} extends LIME \cite{ribeiro2016should} to generate the local explanation of the GNN model on node classification tasks. For an input node, it first constructs a local dataset that consists of the node features of its K-hop neighbor nodes and their predictions, where K depends on the number of GNN layers. Then, it employs an interpretable model, namely  Hilbert-Schmidt Independence Criterion Lasso (HSIC Lasso) \cite{yamada2014high}, to fit the local dataset. The node features are associated with the weights of HSIC kernels, which indicate the importance of the corresponding node features on the outputs of HSIC Lasso. Since HISC Lasso locally fits the output space of the graph model, the important node features are selected as the explanation. The GraphLIME is model-agnostic, and thus can be adapted to interpret any GNN models. However, it has several limitations. First, the HSIC Lasso model is trained to generate customized explanation for each node, which is computationally inefficient. Second, the choices of the interpretable model also influences the produced explanation.

\textbf{RelEx \cite{zhang2021relex}} also focuses on explaining the GNN model on node-level tasks. First, it employs the BFS scheme to generate the perturbed samples of the K-hop subgraph and construct a local dataset like graphLIME. 
Then, it minimizes the disagreement between a  surrogate model and the graph model on the local dataset. Notice that the surrogate model in RelEX is also a graph model instead of an simple and interpretable model for the expressive power. 
Finally, it identifies the important node features of the surrogate model by optimizing a learnable mask, which is used to explain the GNN. Similar to GraphLIME, RelEx is time-consuming to generate the explanation because the local dataset and the surrogate model are customized for each input instance. Moreover, the uninterpretability of the chosen surrogate model also leads to the reduced transparency in explanation.

\textbf{PGM-Explainer \cite{vu2020pgm}} employs a probabilistic graphical model to explain the GNNs. It perturbs the node features in the K-hop subgraph of a central node to construct the local dataset. A random variable is introduced to record whether the node feature is perturbed. The perturbation scheme is to replace the node feature with its mean value for a pre-defined probability. Next, an explainable Bayesian network is learnt to fit the local dataset and generation the explanation. Since it is intractable to learn a Bayesian network, the Grow-Shrink algorithm is used to prune the local dataset. 
Notice that PGM-Explainer can interpret GNNs on both node and graph classification tasks.

\textbf{GraphSVX \cite{duval2021graphsvx}}  introduces the game theory to exploit the node features and graph structures in order to explain the GNN model on both node and graph classification tasks. For an input graph, it produces the perturbed samples by jointly applying the node masks and feature masks to the input. Then, a graph generator maps the perturbed samples to the input domain and constructs the perturbed dataset. An interpretable Weighted Linear Regression (WLR) model is employed to fit the GNN model on the perturbed dataset. The coefficients of the WLR model measure the importance of the corresponding nodes and the node features. 
Moreover, the coefficients are also interpreted as an extension of Shapley value to the graph-structured data from a decomposition perspective. Apart from the explanation tasks, GraphSVX can be adopted to identify the noisy features in real-world graph datasets.

\subsubsection{Generation-based Methods}

\textbf{XGNN \cite{yuan2020xgnn}} first explains the GNN model from a global perspective. Instead of generating customized explanation for an input instance, it generates essential graph patterns that are maximally relevant to a certain class. To do that, it employs a graph generator to refine the graph patterns from the initial states based on reinforcement learning (RL). At each step, XGNN predicts how to add an edge to extend the generated graph for update. The whole framework is optimized by maximizing the reward using policy gradient.
Moreover, several graph rules are introduced to generate valid graphs. Unlike other methods, XGNN provides a global understanding of the GNN model with the class-relevant graph patterns, which are more intuitive. XGNN is also able to be easily generalized to any GNN-based models for interpretation since it do not depend on the certain GNN architecture. However, it is less precise than the instance-level methods \cite{shan2021reinforcement}.

\textbf{RG-Explainer \cite{shan2021reinforcement}} also employs reinforcement learning (RL) to generate the explanations of GNNs’ predictions. It emphasizes the connected nature of the explanatory subgraph since a connected substructure captures the interactions among nodes and edges. 
The key of RG-Explainer is to maximize the mutual information between the input prediction and the label distribution of generated explanatory subgraphs. First, it selects the starting point for exploration. For node-level tasks, the starting point is naturally choosed as the node instance. For graph-level tasks, RG-Explainer seeks the starting point with a learnable multi-layer perceptron (MLP). Then it iteratively adds nodes to the current subgraph until the stopping criteria satisfies. Compared with other RL-based methods, RG-Explainer has superior generalization power to infer the explanations of unseen instances. 

\textbf{GEM \cite{lin2021generative}} formulates the GNN explanation problem as a causal learning task. For and input instance, it evaluates the individual casual effect of each edge and construct the casual subgraph for the prediction. Then, GEM employs a graph auto-encoder to map the input instance to the causal subgraph. Thus, the graph auto-encoder can generalize to unseen graph instance and produce the causal explanation. GEM is model-agnostic, and can be adapted to both node-level and graph-level tasks. Moreover, it enjoys accurate explanation from a causal perspective with low computational complexity. 

\subsubsection{Perturbation-based Methods}

\textbf{GNNExplainer \cite{ying2019gnnexplainer}} explains the prediction of GNN by learning a soft mask for edges and node features, which can be interpreted as sub-graphs and sub-features that are highly related to the prediction. Specifically, GNNExplainer randomly initializes the soft mask as a trainable parameter, and then exploits it to perturb the original adjacency matrix and node features via element-wise multiplications. The masks are optimized by maximizing the mutual information between the predictions of the original graph and that of the perturbed graph. To ensure the sparsity (discrete) of masks, GNNExplainer also introduces different regularization terms, such as element-wise entropy. However, such regularizers cannot guarantee obtaining a strictly discrete mask. Moreover, the masks are optimized for each given graph individually, which is impractical and limited to a local explanation.

\textbf{PGExplainer \cite{luo2020parameterized}} proposes to explain the predictions via edge masks. Given a graph as input, it first obtains the embedding of each edge by concatenating its nodes embedding. 
Then the edge predictor uses the corresponding edge embedding to learn a Bernoulli distribution that represents the edge should be masked. To train in an end-to-end manner, the Gumbel-Softmax trick is applied to approximate the Bernoulli distribution. Finally, the predictor is trained by maximizing the mutual information between the prediction of original graphs and that of masked graphs. Note that, even though the Gumbel-Softmax trick is exploited, PGExplainer still learns a soft edge mask. However, since the all edges in the dataset share the same predictor, the explanations can provide a global understanding of the trained GNNs.

\textbf{GraphMask \cite{schlichtkrull2020interpreting}} explains the trained GNNs by masking the edges for each GNN layer. Similar to the PGExplainer, GraphMask also learns an edge predictor to identify which edges should drop without changing the original graph prediction. However, unlike PGExplainer, GraphMask does not change the topology structure of graphs. Instead, it drops the edge by replacing its embedding with a baseline, which is a learnable vector parameter with the same dimension as node embeddings. Moreover, the straight-through Gumbel-Softmax trick is applied to obtain strictly discrete masks. By maximizing the mutual information as PGExplainer using the whole dataset, GraphMask could learn a global understanding of the trained GNNs.

\textbf{ReFine \cite{wang2021towards}} proposes to generate multi-grained explanations from global and local perspectives. Like previous explainers, ReFine learns discrete mask predictors by maximizing the mutual information of predictions between original and perturbed graphs. To obtain multi-grained explanations, the mask predictor is trained through two-stage training (i.e. pre-training and fine-tuning). The pre-training phase maintains global information by training the predictor over different graphs and distilling the class-level knowledge via contrastive learning. Then the fine-tuning phase further adapts the global explanations in local context by fine-tuning the predictor over a specific instance. Although ReFine improves the quality of explanatory subgraphs, it is still unclear why the pretraining-finetuning could generate multi-grained explanations.

\textbf{Zorro \cite{funke2021zorro}} understands the decisions of GNNs by identifying important input nodes and node features. It exploits a greedy algorithm to select nodes or node features step by step. At each step, Zorro selects the one with the highest fidelity score, which measures how well the new prediction matches the original one if the original graph is perturbed by replacing the unselected nodes and features with random noise values. Obviously, such a greedy algorithm is time-consuming. Besides, it may lead to local explanations since the selected processes are considered for each graph individually.

\textbf{SubgraphX \cite{yuan2021explainability}} explains the trained GNNs by generating subgraphs that are highly correlated with model predictions. Specifically, it exploits the Monte Carlo Tree Search (MCTS) algorithm to explore different subgraphs via node pruning and select the most important one as the explanation. The reward of MCTS is calculated via Shapley value, which measures the contribution of subgraphs to the prediction. To reduce complexity, the computation of Shapley value only involves the interactions within message passing ranges. Even though SubgraphX does not learn a mask explicitly, the node pruning actions in MCTS can be viewed as an implicit discrete mask. Thanks to the employ of Shapley value, SubgraphX is arguably more human-intelligible that other perturbation-based methods. However, the computational cost is undesirably expensive, since it requires exploring different subgraphs with the MCTS algorithm.

\subsubsection{Counterfactual-based Methods}

\textbf{CF-GNNExplainer \cite{lucic2021cf}} first studies the counterfactual explanations for GNN models. Instead of discovering the explanation which accounts for the prediction, the counterfactual approach seeks for the minimal perturbation on the input instance to change the original prediction. Hence, CF-GNNExplainer recognizes a minimal subgraph of the input graph, if removed, would lead to a different prediction. 
Specifically, given an input graph, CF-GNNExplainer iteratively removes edges from the adjacency matrix based on matrix sparsification techniques until the prediction of the input changes. Thus, the obtained counterfactual subgraph is minimal and accurate to explain the prediction changes. Compared with other explanation methods, CF-GNNExplainer interprets the GNN models in terms of the prediction dynamics, leading to more robust explanation for the noisy input \cite{bajaj2021robust}. However, it suffers from the faithfuness issue as the counterfactual subgraph does not necessarily account for the original prediction of the input instance \cite{yu2021towards}.

\textbf{RCExplainer \cite{bajaj2021robust}} enhances the counterfactual explanation to be robust to the input noise. As the GNN models are highly non-convex, their explanations are easily affected by the input noise. Hence, RCExpaliner first models the decision logic of a GNN by the linear-wise decision boundaries. Then, it learns a mask to explore the explanatory subgraph of an input instance. The explanatory subgraph is endowed wit the following properties: 1. faithful and counterfactual to the prediction; 2. sparsity. Moreover, the entropy regularization encourages the learnable mask to be discrete. Compared with CF-GNNExplainer, RCExplainer is more robust to the input noise. Meanwhile, the explanatory subgraph preserves the faithfulness to the original prediction.

\textbf{MEG \cite{numeroso2021meg}} identifies the counterfactual explanation for molecule property prediction. Similar to the prior work, it discovers a minimal substructure of the molecule which would lead to a change in prediction if removed. Differently, it employs a generator to iteratively refine the obtained substructure based on Reinforcement Learning. Moreover, MEG introduces several rules to ensure the chemical validity of the explanatory substructure. 

\textbf{GNN-MOExp \cite{liu2021multi}} studies the joint effect of simulatability and counterfactual relevance on the human understanding of an explanation. The simulatability refers to how an explanation faithfully simulates the prediction of the input instance. And the counterfactual relevant means whether the removal of an explanation leads to a change in prediction. Given an input graph, GNN-MOExp generates the explanatory subgraph by conjecturing the cognition process of human. Specifically, it first evaluates the simulatability of the explanation at the first stage. The explanatory subgraph is send to the next stage if it has a high simulatability value or rejected otherwise. Then, GNN-MoExp examines its counterfactual relevance at the second stage. Finally, the acceptance score of an explanation is obtained by considering both stages. Moreover, GNN-MOExp design a multi-objective optimization (MOO) algorithm to find an explanation which is pareto optimal for multiple objectives. Hence, GNN-MOExp generates more comprehensive and human-intelligible explanation.

\subsection{Self-explainable Methods}
The self-explainable methods embed the intrinsic explanations in the architectures of the GNN models so that they can make predictions and generate the corresponding explanations during the inference time, simultaneously. Generally, they either recognize the predictive substructures of the input graph for the explanation, or induce the evidence of the outputs via regularization.

\textbf{GIB \cite{yu2020graph}} extends the Information Bottleneck \cite{tishby2000information} method to the graph learning and proposes the model-agnostic Graph Information Bottleneck framework to empower the GNN models with self-explainability. Given an input graph, it  first generates the compressed yet informative subgraph (IB-Subgraph) with a graph generator. Then, the GNN model only takes the IB-Subgraph as input and infer the underlying label or property of the original graph. The whole framework is trained in an end-to-end fashion by optimizing the GIB objective, which consists of a compression term and a prediction term. The compression term minimizes the mutual information between the input graph and the IB-Subgraph, and the prediction term maximizes the mutual information between the IB-Subgraph and the original graph label. The GIB method enjoys several strength with the information-theoretical objective. First, the IB-Subgraph is proved to be noise-invariant \cite{yu2021recognizing}. Thus, it improves the robustness of the GNN model on graph classification task by dropping the redundancy and noisy information. Second, the IB-Subgraph naturally serves as an explanation for the GNN's prediction. However, the GIB framework suffers from time-consuming and unstable training process, mostly due to the intractability of mutual information. To minimize the compression term, it introduces a bi-level optimization scheme to estimate the corresponding mutual information and use the estimated value for minimization. The estimation process is time-consuming, and the inaccurate estimated value leads to unstable training process or even degraded performances \cite{yu2021improving}.

\textbf{VGIB \cite{yu2021improving}} reformulates GIB as the graph perturbation and subgraph selection, and achieves efficient and stable optimization process with the improved performance. For the graph perturbation, a learnable noise injection module selectively inject noise into the input graph to obtain a perturbed graph. Then, VGIB encourages the perturbed graph to be informative of the graph property. Hence, only the insignificant substructure are replaced with noise and the crucial subgraph is well preserved. This formulation allows a tractable variational upper bound of the information loss in the perturbed graph. Thus, the VGIB objective is easy to optimize and free from bilevel optimization. VGIB can be generate explanation for both node-level and graph level tasks. Moreover, it be plugged into any GNN models for self-explainability and generate post-hoc explanations for pretrained GNNs as well.

\textbf{DIR \cite{DIR}} employs the causal inference to enable the GNN models with intrinsic interpretability. Since the real-world tasks may encounter data biases and out-of-distribution data, the explanation of GNN models may rely on the short cuts in the graph-structured data. To this end, DIR first disentangles the graph into causal and non-causal subgraphs, which are respectively encoded by the encoder into representations. Then, it treats whether the non-causal subgraph is involved as different environments, and learns the intervention distributions to distill the invariant subgraph of the models. Hence, DIR serves as a probe for the inference while achieves the self-explainable graph model. Moreover, the learnt invariant subgraph demonstrate to be effective in generalization.

\textbf{SEGNN \cite{dai2021towards}} introduces an explainable framework for graph classification based on K-Nearest Neighborhood clustering. Specifically, given an unlabeled node for testing, it searches K-nearest labeled nodes based on the node similarity and local structural similarity. Then, the label for the testing node is chosen as the majority of the K-nearest labeled nodes. The node similarity is computed by comparing the node embeddings, which are encoded by a GNN model. The structured similarity is obtained by the similarity of the edges of their ego graphs. SEGNN provides a heuristic manner to interpret the prediction in common graph tasks. However, its simple architecture limits its applications in more challenging tasks.

\subsection{Datasets and Tools}

The evaluations of graph explanation are non-trivial because annotating the ground-truth explanations require intensive labor and even expertise. First of all, to mitigate the lack of annotations, several synthetic datasets are built by approximating the ground truths of graph explanations through graph motifs \cite{ying2019gnnexplainer}. BA-shapes is a node classification dataset with 4 different kinds of nodes. For each graph, it contains a base graph and a set of 80 five-node house-structured motifs. The base graph is generated by the Barabasi-Albrt (BA) model with the preferential attachment mechanism \cite{albert2002statistical}. Then the node motif is attached to the base graph with random generated edges. Hence there are 4 different classes of nodes, including the top, middle and bottom nodes of the house, and nodes outside the house. Extended from BA-shapes, BA-Community randomly connects two BA-shapes graphs with random edges to generate a new graph. Compared with BA-shapes, the nodes of BA-shapes have 8 different locations and their labels are assigned based on their structural locations and the memberships of BA-shapes graphs. Tree-Cycle synthesizes the graph with a base 8-level balanced binary tree graph and a six-node cycle motif. The cycle motif is attached to random nodes of the base graph. The labels of nodes are 1 when they are located in the base graph. Otherwise, their labels are 0. Similar to Tree-Cycle, the Tree-Grids dataset randomly attaches 3-by-3 nodes grid motifs to the base tree graph rather than the cycle motif. The BA-2Motifs dataset combines BA-shapes and Tree-Cycle. It synthesizes 800 graphs by attaching the house-structural motifs and the five-node cycle motifs to the base BA graph. The graphs are assigned binary labels according to the type of attached motifs. 
The synthetic datasets are employed to evaluate GNNs whether their explanations can capture the graph structures. However, due to the simple synthesis, the relationships between graphs and labels in these datasets are simple and thus lead to simplistic evaluations. 

Besides, GNN explanations are expected to be intuitive and provide understandable visualizations for human. To this end, several sentiment graph datasets such as Graph-SST2 \cite{socher2013recursive}, Graph-SST5 \cite{socher2013recursive}, and Graph-Twitter \cite{dong2014adaptive} are built with text. GNNs use words and phrases to interpret explanation results. These datasets contain text sequences. Each sequence is used to construct graph, whose nodes are embeddings of words and edges are the relationships between different words. (The article has not introduced How to annotate the label information.)
Moreover, several molecular datasets including MUTAG \cite{debnath1991structure}, BBBP \cite{debnath1991structure}, Tox21 \cite{wu2018moleculenet}, QED \cite{yu2021improving}, DRD2 \cite{yu2021improving}, HLM-CLint \cite{yu2021improving} and RLM-CLint \cite{yu2021improving} are proposed for GNN explanations. In these datasets, each molecule is converted to a graph, whose nodes are the atoms and edges are the chemical bonds. The labels of molecules are annotated by chemical experts according to the chemical functionalities or properties of chemical groups.

Along with the datasets for evaluation, several packages such as PYGeometric\footnote{\url{https://github.com/pyg-team/pytorch_geometric}} and DIG\footnote{\url{https://github.com/divelab/DIG}} are now publicly available and can be used to implement the explanation methods for GNNs. There is also an orthogonal survey on the progress of the explanability of GNNs \cite{yuan2020explainability}.
\subsection{Applications}

The Graph Neural Networks (GNN) have been widely adopted to various tasks for the superior performance on learning with graph-structured data. In chemistry, GNNs are employed to analyze the chemical properties of the molecules \cite{ma2020deep,rong2020self} and to facilitate the de novo molecule design \cite{jin2020multi,yu2022structure,chen2020molecule}. For histopathogy, the biological entities such as cells, tissue regions, and patches of histopathology images are embedded into biological graphs \cite{pati2020hact}. Hence, the GNN-based models become prevalent in processing the biological graphs and facilitate clinical decisions \cite{jaume2021histocartography}. Moreover, these explanation methods also provide a deeper understanding of GNN-based reasoning \cite{schlichtkrull2020interpreting, minervini2020learning}.
Accordingly, explanation techniques are extended to accommodate the GNN-based methods with improved explainability and reliability in these fields.

In chemistry, the explanation methods are employed to probe the influential substructures of the molecules. For many molecule-level tasks, scientists are curious about which substructure of an given molecule has the largest impact on the molecule property \cite{rao2021quantitative}. Such substructure, namely the functional group, is a compact and connected subgraph which dominates the chemical functionalities. Several methods resort to Monte-Carlos Tree Search (MCTS) \cite{jin2020multi} and information-theoretic approaches \cite{yu2020graph} to distill the functional groups from the labeled molecules. The obtained functional groups can further facilitate de novo molecule design \cite{jin2020multi,chen2020molecule}. Meanwhile, since GNN models are widely adapted to predict the chemical properties of molecules, some regularisations are introduced to enhance the interpretability of the predictive results \cite{henderson2021improving}. 

Besides, the explanation methods accommodate GNN-based clinical decisions with the improved transparency. Recently, many biological entities, such as cells, tissues and patches of digital pathology images, are modeled as biological graphs to better exploits the their interactions. Hence, GNN-based methods show superior performances on biological graph analysis for their inductive bias on graph-structured data.
A large body of these works focuses on the application of GNNs on cancer subtype classification, such as breast cancer classification \cite{rhee2017hybrid, anand2020histographs, jaume2020towards}, colorectal cancer classification \cite{zhou2019cgc,studer2021classification} and lung cancer classification \cite{li2018graph,adnan2020representation}. However, there is increasing concerns on the reliability of the GNN-based models as they are treated as the black-box. Althogh several explanation methods of GNN are implemented to digital-pathology tasks \cite{jaume2021quantifying,yu2021towards,jaume2020towards}, it still remains challenging to develop the explanation methods which align with the domain knowledge of the clinical practitioners.

The explanation methods also uncover the rationale behind the prediction in GNN-based reasoning by underlining the essential edges, paths and random walks in the knowledge graphs. These methods mainly include interpretable question/query-answering \cite{schlichtkrull2020interpreting, minervini2020learning} and explainable neural reasoning \cite{liu2021neural,shen2018m}. As the applications of explanation methods of GNN is still an emerging area, there are many opportunities as well as challenges in other high-stake scenarios such as finance, medicine, legal system and etc.
\section{Privacy}
\label{sec:privacy}


\subsection{Overview and Taxonomy}
The recent successes of graph learning are typically built upon large scale graph data. However, in practice, various data regulations (e.g., GDPR \cite{} and CCPA \cite{}) and privacy concerns from public have become a major concern that prevents organizations from sharing their own graph data with others. Here we take building e-commerce recommendation systems as an example. In this case, all users and items can form a large graph wherein a link indicates whether a user is interested to a specific item. In this case, the users' features typically distribute over different organizations. For example, the deposit and loans of a user is always stored in a financial organization (e.g., WeBank and AntGroup) while the shopping history is always stored in the e-commerce organizations (e.g., JD and Alibaba). If we further take the social connections between users into consider, this data is typically stored in social network departments such as Wechat. As we can see from above example, the user-item graph is demonstrated as a large  heterogeneous graph wherein subgraphs are held by different organizations or departments, i.e, ``data isolated island''. This issue presents serious challenges for applying graph learning to real-world scenarios. 

To solve this issue, numerous efforts have been made to build privacy-preserving graph learning (PPGL) \cite{peng2021differentially,igamberdiev2021privacy,chen2021fedgl,fedgnn_framework,asfgnn} which aims to build graph learning framework while protecting users' data privacy. In this survey, we categorize previous research work of PPGL into three directions, namely, federated graph learning, privacy inference attack, and private graph learning. Specifically, federated graph learning
(FGL) aims to provide a general distributed learning paradigm enabling multiple organizations to jointly train a global model without sharing their own raw data. While FGL has provided the protection for raw data, most work still need to share intermediate data such as gradients or hidden features and these data can also contain some sensitive information. This information can be reconstructed by some specific privacy inference attacks which is the second part of this section. Further, we review some recent private graph learning work which aims to employ cryptographic techniques to prevent above attacks.

\subsection{Federated Graph Learning} \label{subsec:fedgraph}
A recent paper~\cite{fedgl_position} categorizes previous FGL methods into four types based on how graph data are distributed among devices. 
In this survey, we follow their categorization 
and introduce some typical FGL methods in including their limitation and potential improvements in each direction.

\nosection{Inter-graph federated learning}
Inter-graph federated learning is similar to the setting proposed in the original paper~\cite{fl_original}. In this setting, each sample held by clients is a complete graph and the target task is  graph-level prediction. A typical scenario under this setting is federated drug development. An important application in this scenario is molecular property prediction where each  pharmaceutical company holds some molecular graphs and associated labels. The goal of this application is to jointly train a global prediction model without exchanging confidential datasets held by different organizations. Inter-graph federated learning can be naturally implemented by conventional FL methods introduced in the above section, such as FedAvg, FedProx, and FedDyn. FedGNN~\cite{fedgnn_framework} has provided an implemented example for this setting on a number of drug-related tasks.

\nosection{Horizontal Intra-graph federated learning}
Intra-graph federated learning refers to each client holds a part of the entire graph (i.e., a sub-graph). 
Intra-graph FL can be further divided into horizontal and vertical FGL based on the prior work \cite{yang_fl}.  
\textbf{For the horizontal setting, subgraphs held by different clients share the same features but different node IDs.} 
Many B2C (Business-to-customer) scenarios belong to this setting. A typical example is user behavior modeling in online social platform,  where each terminal user has a local social network and server wants to train a model to describe the user's behavior (e.g., fraud user detection and recommendation). In this case, horizontal intra-graph FL enables to train a powerful global model by leveraging information from all terminal users without violating their privacy \cite{wang2020graphfl,chen2021fedgl,xie2021federated}. Here, we introduce a prior work \cite{fedgnn_rec} focusing on federated recommendation modeling under this horizontal setting. This work treats user-item interactions as a heterogeneous graph where each item and user are represented as graph nodes. The user-item edge represents that user has rated the item. The rate is denoted as the edge weight. The user-user edge represents these two users have first-order social relationship. 
In the vertical FL setting, each client in this application only holds a subgraph formed by the corresponding user and his interested items. FedGNN~\cite{fedgnn_rec} is presented to train a global GNN model that exploits high-order user-item interactions without privacy leakage. The framework consists of two parts, i.e., private model update generator and private user-item graph expansion. The first part is to generate model update exposed to the sever without users' information leakage. In this case, information leakages include two parts: (1) non-zero embedding gradients/updates can expose which items have been rated by the corresponding user. (2) parameter gradients/updates may contain users' information which can be recovered via carefully devised attack methods. To prevent the first type of leakage, this work introduces pseudo
interacted item sampling. Specifically, they sample  items that the
user has not interacted with and randomly generate their gradients using a Gaussian distribution with the same mean and
co-variance values with the real item embedding gradients.  Then all the real and pseudo gradients are sent to the server. To prevent the second type of leakage, the author presents to inject noise into the gradient to satisfy local differential privacy. The second block of this work is private user-item graph expansion for leveraging information from other users without exposing their information. Specifically, the server first generates a public key  and sends it to each clients. Then each client encrypts the interacted item ids and embeddings and sends the encrypted data  to a third party. For each client, the third party find its neighbors based on the encrypted data and sends item ids to the client. The client can update its local model based on the information from the neighbors. 

\nosection{Vertical Intra-graph federated learning}
\textbf{In contrast to horizontal setting, in the vertical setting, each client shares the same node IDs but different node features.} This setting wildly exists in many B2B (business-to-business) scenarios. A typical example in real industry is federated credit modeling between financial and e-commerce platform. In this case, these two platforms share the same activate user groups. For node features, financial platform may have users' deposit and loans and e-commerce platform has users' shopping history and interested items. The goal is to train a credit model based on data from both two platforms without violating data regulations. Chen et al. \cite{vertical_gnn} proposes a general framework for this setting combining federated and split learning \cite{split_learning}. Specifically, this work splits GNN computation into two parts, i.e., private and non-private parts. Here, private part is conducted over private users' private data  and the non-private part is conducted over some intermediate data such as the hidden embeddings. To fully release the power of users' data across different platforms, this work employs secure
multi-party computation (MPC) technique to enable all platforms to jointly compute initial node embeddings without sharing their private data. Once the embeddings are obtained, each platform can perform local message passing over the initial node embeddings based on the node connection information they hold to obtain local hidden embeddings. Then these hidden embeddings are sent to the server and the server aggregates these local embeddings to obtain the final global node embeddings. At last, the platform who holds label information can perform prediction using the global node embeddings. To further protect users' privacy, this work proposes to inject noise into the local embedding computation to achieve differential privacy~\cite{dp_dwork}.

\nosection{Graph-structured federated learning}
In graph-structure federated learning, the topology of the latent complete graph is distributed over different organizations. In this setting, each client holds different relationships among different nodes \footnote{Here, we assume that all clients share the same node space and put the focus on graph topology}. This setting  can be seen as a special case of vertical intra-graph federated learning with particular focus on the graph topology instead of node features. One typical
application is federated traffic flow prediction~\cite{fedgl_position} where the goal of the central server is to train a GNN model leveraging the relationships among different edge devices distributed in different geographic positions.
\subsection{Privacy Inference Attack}


In the era of big data, an extensive collection of real-world applications can be represented as graphs, such as social relation graphs, bipartite recommendation graphs, and molecule graphs. Due to graphs’ discrete nature, it is challenging for deep learning to utilize them directly. To address this issue, graph representation learning techniques like graph embedding and graph neural networks have been proposed to project nodes, edges, or even subgraphs into low-dimensional vector representations. These representation learning modules are either part of an end-to-end model, or an individual model that produces embeddings for downstream tasks. 
In the past decade, the success of these graph representation learning techniques has been demonstrated in various real-world tasks, such as online recommendation, anomaly detection, drug development, etc. Despite much success, the associated privacy concerns remain unanswered for a long time. This section provides an overview of the privacy attacks to graph representation learning techniques, followed by a comprehensive review of attacks proposed in recent works.

\subsubsection{Overview}

Recent literature has revealed two major categories of privacy threats in graph learning techniques, i.e., membership inference attacks (MIA) and model extraction attacks (MEA). Both threats happen in the test phase of a trained model. MIA aims to determine if a target graph element (e.g., a node or an edge) is within the victim model’s training set, while the goal of MEA is to reconstruct a surrogate model with similar performance as the victim model. In a nutshell, MIA and MEA reveal privacy issues of graph learning techniques on data-level and model-level, respectively. 

\subsubsection{Membership Inference Attacks.}

\begin{table}[h]
    \centering
    \caption{Threat Models of MIAs.}
    \label{tab:MIA}
    \begin{tabular}{c|c|c}
    \toprule
    \diagbox{Threats}{Victims} & Graph Embeddings & GNNs \\
    \hline
    Attribute-Level & \cite{DBLP:conf/mobiquitous/DudduBS20} & \cite{DBLP:journals/corr/abs-2110-02631}  \\
    \hline
    Node-Level & \cite{DBLP:conf/mobiquitous/DudduBS20, DBLP:journals/corr/abs-2104-08273} & \cite{DBLP:journals/corr/abs-2102-05429,DBLP:journals/corr/abs-2101-06570} \\
    \hline
    Edge-Level & \cite{DBLP:journals/corr/abs-1912-10979} & \cite{DBLP:conf/uss/HeJ0G021}  \\
    \hline
    Graph-Level & \cite{DBLP:conf/mobiquitous/DudduBS20} & 
    \cite{DBLP:journals/corr/abs-2110-02631}  \\
    \hline
    Blackbox & \cite{DBLP:conf/mobiquitous/DudduBS20, DBLP:journals/corr/abs-2104-08273} &  
    \cite{DBLP:conf/uss/HeJ0G021,DBLP:journals/corr/abs-2102-05429,DBLP:journals/corr/abs-2101-06570, DBLP:journals/corr/abs-2110-02631}  \\
    \hline
    Whitebox & \cite{DBLP:journals/corr/abs-1912-10979, DBLP:conf/mobiquitous/DudduBS20}  &  \\
    \bottomrule
    \end{tabular}
\end{table}
In MIA, an adversary is given the target element (e.g., the nodes, edges), information about the training graph (e.g., local topology), intermediate outputs of the victim model (e.g., node embeddings), and black/white-box access to the victim model. We offer a summary of the MIAs mentioned in this section in Table \ref{tab:MIA}.

Early MIA works focus on analyzing the privacy issues of unsupervised graph embedding techniques~\cite{DBLP:conf/mobiquitous/DudduBS20,DBLP:journals/corr/abs-1912-10979}. As a pioneer work \cite{DBLP:journals/corr/abs-1912-10979}, the authors investigate the possibility of recovering edges for a node using the embedding of the nodes in the remained graph and their topological structure. The paper assumes a white-box setting, in which the embedding algorithm and its parameters are known to the attackers. The proposed multi-phase approach first randomly samples a collection of nodes and then learns node embeddings on the graphs with and without every single node. Thus, the adversary can use these `positive' and `negative' samples to train a classifier to determine whether a specific node is the removed node's neighbor. Thus, the edges of the removed node are recovered.

Later in In \cite{DBLP:conf/mobiquitous/DudduBS20}, the authors consider a wide spectrum of attack including node inference attack, attribute inference attack and graph reconstruction attack. For the node inference attack, the paper considers both white-box (i.e., the adversary has access to the output for intermediate layers) and black-box setting (i.e., the adversary merely exploits the prediction scores in the node classification task).  
For the white-box setting, \cite{DBLP:conf/mobiquitous/DudduBS20} propose to train an autoregressive encoder-decoder network that minimizes the reconstruction loss of the target node's embedding.
For the black-box setting, \cite{DBLP:conf/mobiquitous/DudduBS20} proposes two different approaches, i.e., the shadow attack and the confidence attack. Both attacks are based on the intuition that member and non-member nodes' confidence distributions are significantly different. 
For the attribute inference attack, the paper tries to infer nodes' attributes using the released node embeddings. This is done by training an supervised model that takes node embeddings as inputs to predict sensitive, hidden attributes.
For the graph reconstruction attack, the attacker is also given access to the node embeddings of a subgraph. To accomplish the reconstruction task, the adversary samples additional graphs that have the same distribution as the target graph. The additional graphs are used to train an encoder-decoder model that reconstructs a graph from its publicly released node embeddings. With the trained encoder-decoder model, the attacker can reconstruct the target graph.

Apart from graph embedding, there are also some works that expand MIA to more complicated graph neural networks (GNNs). 
He et al. \cite{DBLP:journals/corr/abs-2102-05429} performed the first systemical study on node-level membership inference attacks against GNNs, which aims to infer whether a given node $v$ is used to train a target GNN model or not. The paper assumes a black-box setting, i.e., the adversary can merely query the target model to get predictions. Under such a setting, the paper proposes a multi-phase approach that includes shadow model training, attack model training, and membership inference. The shadow model, which replicates the behavior of the target model, is trained on an additional graph that follows the same data distribution as the target model’s training set. The attack model is trained to predict whether a node is from a specific graph. Specifically, the adversary first queries the shadow model with 0-hop queries (the target node and a self-loop) and 2-hop queries (target node's 2-hop subgraphs) of nodes to get the prediction vector. The prediction vector is then fed into the attack model A to obtain the membership prediction. Thus, the membership attack is accomplished.
The concurrent work \cite{DBLP:journals/corr/abs-2101-06570} also conducts node membership inference attack on GNNs under the black-box setting. The proposed attack is similar to  \cite{DBLP:journals/corr/abs-2102-05429} and the major difference is that \cite{DBLP:journals/corr/abs-2101-06570} proposes to use target node's $L$-hop neighbors to query the shadow model and get prediction vectors. Recently work \cite{DBLP:journals/corr/abs-2104-08273} evaluates the privacy threats on heterogeneous knowledge graphs. The paper proposes three types of membership inference attacks: transfer attacks (TAs), prediction loss-based attacks (PLAs), and prediction correctness-based attacks (PCAs). Like \cite{DBLP:journals/corr/abs-2101-06570} this paper also assumes that the adversary has black-box access to the target KGE model and can build a shadow dataset.

In addition to node-level membership attacks, some works also focus on conducting graph MIA on edge and graph level. 
In \cite{DBLP:conf/uss/HeJ0G021}, the authors investigate the threat of edge membership inference attack, whose goal is to infer whether a given pair of nodes $u$ and $v$ are connected in the target graph. The authors assume the adversary can only obtain nodes' posteriors by querying the target model, i.e., black-box setting. Also, the adversary knows the target training set's nodes' attributes, part of the target training graph and an additional shadow dataset. The paper introduces eight attacks and the intuition behind these attacks is that \textit{two nodes are more likely to be linked if they share more similar attributes and/or predictions from the target GNN model.}
Based on such an intuition, the proposed attacks transfer the knowledge from the shadow dataset to the target dataset to launch the attack. \cite{DBLP:journals/corr/abs-2110-02631} further expand the membership attack to attribute-level and graph-level. It considers the scenario that the victim GNN releases the learned graph embeddings (aggregation of node embeddings) to third parties. The adversary can then rely on the released graph embeddings, and the black-box access to the target victim model to infer the sensitive information on the GNN's train graph. The sensitive information includes graph properties (properties inference attack), graph topology (graph reconstruction attacks), and subgraphs (subgraph memberships inference attack).
The attack approach is similar to \cite{DBLP:journals/corr/abs-2102-05429} and the major difference is that the authors design different attack models and training tasks for the aforementioned membership attacks.

\subsubsection{Model Extraction Attacks}

Apart from the data-level privacy threat, i.e., membership inference attack, graph learning models are also vulnerable to model-level threats, i.e., model extractions attacks (MEA). MEA aims to construct a model that achieves similar performance to the target model (e.g., similar overall performance or similar prediction distributions). The first MEA approach against GNN models is proposed in \cite{DBLP:journals/corr/abs-2010-12751}. In \cite{DBLP:journals/corr/abs-2010-12751}, the authors first formulate multiple threat models in which attackers are granted different capabilities (i.e., whether the attacker can access node attributes, connections, and the shadow dataset). Then the authors propose attack approaches that utilize the capability in each threat model to launch the attacks. The intuition behind the attack is to reconstruct the missing nodes attributes (or edges) based on the existing edges (or attributes) and the auxiliary shadow graph. Then a surrogate model is trained on the reconstructed graph to duplicate the behavior of the target model.

\subsection{Private Graph Learning}

Although several federated graph learning methods have been studied, as presented in Section \ref{subsec:fedgraph}, the privacy issues in them have also attracted a lot of attention recently. 
For example, the leak of gradient or hidden embedding during federated learning may cause kinds of attacks, e.g., model inversion attack on GNN models \cite{zhang2021graphmi}. 

Existing work has proposed different solutions toward private graph learning. 
Technically, we divide them into two types, i.e., differential privacy (DP) based and secure multi-party computation (MPC) based. 
The former incorporates DP to build private graph learning models with formal privacy guarantees, while the later leverages MPC techniques, e.g., secret sharing, to build provable-secure graph learning models. 
For example, the authors in \cite{sajadmanesh2021locally} develop a privacy-preserving, architecture-agnostic GNN learning algorithm based on local DP, and the authors in \cite{peng2021differentially,igamberdiev2021privacy} propose a differentially-private gradient-based training paradigm for GNN. 
The authors in \cite{chen2020survey} propose a secure multi-party computation (MPC) based private graph learning strategy. 
In it, the the forward computation of graph representation learning is divided into three steps, i.e., secure initial node embedding generation, secure embedding propagation, and secure loss computation. The linear computations in each step could be done directly by leveraging MPC techniques, e.g., secret sharing, while the non-linear computations could be approximated by high-order polynomials. 
DP based private graph learning has high efficiency, however, there is a trade-off between privacy and utility. 
In contrast, MPC based private graph learning can achieve comparable performance with plaintext graph learning models, but it comes with serious efficiency cost. 

\section{Conclusion}
\label{sec:conclusion}
Deep graph learning (DGL) are widely studied recently but the trustworthiness of DGL models is still less explored. This survey provides a systematic and comprehensive overview of trustworthy graph learning (TwGL) from three fundamental aspects, i.e., reliability, explainability and privacy. Specifically, we first introduce the reliability of DGL against three threats, including inherent noise, distribution shift and adversarial attacks. Then we provide detailed review on graph explainability techniques and corresponding applications. We followed by giving research overview on privacy protection on DGL from federated graph learning, privacy inference attack and private enhancing techniques. Finally, we highlight several challenges and future directions which hold promising opportunities.
As an emerging field, TwGL research is starting to gain significant interest and development. We hope our work can shed a light for  researchers and practitioners to better understand the trustworthiness of DGL models, and potentially inspire additional interest and work on this promising research direction.

\bibliographystyle{plain}
\bibliography{ref,reliability/reliability.bib}
\end{document}